\documentclass[12pt, letterpaper]{article}

\usepackage[margin=1in]{geometry}
\usepackage{setspace}
\usepackage{times}
\usepackage[utf8]{inputenc}
\usepackage[T1]{fontenc}
\usepackage{xcolor}
\usepackage{microtype}
\usepackage[font=small,labelfont=bf]{caption}
\usepackage{cite}
\usepackage{amsmath,amssymb,amsfonts,amsthm}
\usepackage{algorithm}
\usepackage{algorithmic}
\usepackage{graphicx}
\usepackage{booktabs}
\usepackage{multirow}
\usepackage{tikz}
\usetikzlibrary{shapes.geometric, arrows.meta, positioning, calc, shadows, fit}
\usepackage{url}
\usepackage{hyperref}

\newtheorem{definition}{Definition}
\newtheorem{proposition}{Proposition}

\newtheorem{corollary}{Corollary}

\newtheorem{remark}{Remark}
\newtheorem{assumption}{Assumption}

\newcommand{\Senv}{\mathcal{S}_{\text{env}}}
\newcommand{\Swork}{\mathcal{S}_{\text{workflow}}}
\newcommand{\Agen}{a_{\text{gen}}}
\newcommand{\Rmax}{R_{\max}}
\newcommand{\emax}{e_{\max}}
\newcommand{\rpatience}{r_{\text{patience}}}

\begin{document}

\begin{titlepage}
    \centering
    \vspace*{1.5in}

    {\Huge \bfseries The Dual-State Architecture\\for Reliable LLM Agents \par}

    \vspace{1.5in}

    {\Large \textbf{Matthew Thompson} \par}
    \vspace{0.2in}
    {\large \textit{Independent Researcher} \par}
    {\large matthew@agentia.tech \par}
    {\large ORCID: 0009-0007-0846-0369 \par}

    \vspace{1.0in}

    {\large \today \par}

    \vfill

    {\footnotesize \textit{Preprint submitted to arXiv.\\
    This paper extends and supersedes the execution framework presented in arXiv:2512.20660.\\
    New contributions: tri-state guard semantics (\S3), three-level recovery hierarchy (\S4),\\
    and SWE-Bench boundary analysis (\S5.4).}}
\end{titlepage}

\begin{abstract}
Large Language Models deployed as code generation agents exhibit stochastic behavior incompatible with the deterministic guarantees required by software engineering. We formalize the \emph{Dual-State Action Pair} (DSAP), an execution primitive that couples stochastic generation with deterministic post-condition verification. Guard functions act as sensing actions that project opaque LLM outputs onto observable workflow state, enabling a dual-state decomposition: finite, deterministic $\Swork$ paired with infinite, stochastic $\Senv$. We prove that for $\epsilon$-capable generators, failure probability $P(\text{fail}) \leq (1{-}\epsilon)^{\Rmax} \to 0$. To prevent naive $O(R^K)$ retry explosion across multi-step workflows, we introduce a three-level recovery hierarchy: context refinement (retry within step), informed backtracking (stagnation detection with cascade invalidation and context injection to upstream steps), and human escalation. Experimental validation across 13 LLMs (1.3B--15B parameters) on three diagnostic probes demonstrates reliability gains of up to 66 percentage points at 1.2--2.1$\times$ baseline cost. Recovery mechanism evaluation on 99 SWE-Bench Pro instance-arm pairs (Qwen3-Coder-Next) demonstrates 100\% context injection effectiveness (upstream output changed in all 71 escalation events) with step-specific recovery asymmetry---37.5\% for test generation vs.\ 0\% for patch generation---and 0\% end-to-end patch production, establishing the boundary between execution architecture and plan synthesis: execution recovery is necessary but not sufficient for autonomous software engineering.

\end{abstract}


\section{Introduction}

\subsection{The Problem}

Software engineering demands deterministic guarantees---syntax correctness, type safety, API contract adherence---yet Large Language Models (LLMs) operate as probabilistic approximate reasoners. Current approaches to LLM-based code generation primarily rely on prompt engineering~\cite{wei2022chain} or internal reasoning chains~\cite{yao2022react} to induce reliability. These locate the control loop \emph{inside} the stochastic generation window, where decision-making is subject to the same hallucination modes as content generation.

The fundamental impedance mismatch is representational: classical planning assumes deterministic action outcomes~\cite{ghallab2004automated}, while LLM inference produces stochastic outputs with no guarantee of satisfying post-conditions. Empirical work confirms that LLMs fail to verify their own solutions~\cite{valmeekam2024planning}, establishing that the generation of an artifact and the verification of its correctness are fundamentally distinct operations that cannot both be delegated to a stochastic process.

\subsection{Our Approach}

This work builds on the \emph{Dual-State Architecture} proposed in~\cite{thompson2025managing}, which separates deterministic control flow from stochastic content generation. The central mechanism is the \textbf{Dual-State Action Pair} (DSAP): an indivisible coupling of a stochastic generator with a deterministic \textbf{Guard Function} (Figure~\ref{fig:atomic_pair}). Unlike pre-condition gates in classical goal-based agent frameworks~\cite{shen2003goal}, our guards are \emph{post-condition validators}---sensing actions that project the opaque output of an LLM onto an observable workflow state. The system state space decomposes into $\mathcal{S} = \Swork \times \Senv$ (Definition~\ref{def:state_space}), where $\Swork$ is a finite, deterministic space tracking guard satisfaction and $\Senv$ is the infinite, stochastic space of generated artifacts and accumulated context.

This paper addresses \emph{execution} within human-authored workflows. The planning problem is formally defined (Definition~\ref{def:planning}), but plan synthesis is not automated: the contribution is an execution atom and recovery mechanism, not a planner.

\subsection{Contributions}

\begin{enumerate}
    \item \textbf{C1: The Dual-State Action Pair (DSAP).} A formal execution primitive coupling stochastic generation with deterministic post-condition verification. We prove asymptotic soundness (Proposition~\ref{prop:soundness}) and derive reliability bounds (Corollary~\ref{cor:reliability}).
    \item \textbf{C2: Three-Level Recovery Hierarchy} (Figure~\ref{fig:recovery}). Context refinement (Level~1), informed backtracking with stagnation detection, cascade invalidation, and context injection (Level~2), and human escalation (Level~3). The hierarchy prevents naive $O(R^K)$ retry explosion (Proposition~\ref{prop:ork}).
    \item \textbf{C3: Experimental Validation with Boundary Finding.} DSAP effectiveness across 13 LLMs with up to 66pp improvement; recovery mechanism behavioral validation on SWE-Bench; and boundary finding establishing that execution recovery cannot substitute for plan synthesis.
\end{enumerate}

\section{Related Work}

\subsection{Classical Planning and PDDL}

Classical planning~\cite{ghallab2004automated} relies on the assumption that if an action's preconditions are met, the agent produces the specified effect deterministically. LLM inference violates this contract entirely. Table~\ref{tab:pddl_mismatch} summarizes the representational mismatch.

\begin{table}[t]
\centering
\caption{The PDDL mismatch: LLM inference violates classical planning assumptions.}
\label{tab:pddl_mismatch}
\begin{tabular}{l l l}
\toprule
\textbf{PDDL Component} & \textbf{Classical Assumption} & \textbf{LLM Reality} \\
\midrule
Initial State & Ground atoms (inspectable) & Prompt embedding (opaque) \\
Action Schema & Preconditions $\to$ Effects & No precondition gating \\
Result Function & Deterministic ($s' \leftarrow s \cup \text{ADD}$) & Stochastic ($P(\text{token}|\text{ctx})$) \\
Goal State & Entailment test ($s \models G$) & Implicit / none \\
\bottomrule
\end{tabular}
\end{table}

\subsection{LLM Control Architectures}

Approaches to LLM control fall into two categories. \emph{External control} architectures---including HTN planning~\cite{erol1994htn} and LLM-Modulo~\cite{kambhampati2024llms}---maintain deterministic oversight outside the model, but rely on explicit causal models difficult to extract from black-box LLMs. \emph{Internal control} architectures---Chain-of-Thought~\cite{wei2022chain} and ReAct~\cite{yao2022react}---locate reasoning inside the generation window, achieving flexibility at the cost of probabilistic control flow.

Our framework synthesizes these perspectives: external symbolic guards enforce convergence of internal generative processes, without requiring explicit causal models of the generator's internals. This aligns with the agent-environment boundary~\cite{sutton2018reinforcement,russell2020artificial}: since the LLM's weights cannot be modified during inference, it resides in the environment, with the agent function mapping percepts (generation outputs) to actions (verification decisions).

\subsection{Goal-Net and CSGM}

The Composite State Goal Model (CSGM)~\cite{shen2003goal} introduces explicit guard functions within a Petri net representation. However, CSGM guards are \emph{pre-condition gates}: they determine whether a transition \emph{should fire}, assuming that if permission is granted, the agent executes deterministically. This does not address the stochastic generation problem; it merely gates it. Since 1-safe Petri nets and state-transition systems are computationally equivalent~\cite{rintanen2009planning}, we adopt the state-transition representation, which allows guards to evaluate arbitrary semantic properties of generated artifacts.

Table~\ref{tab:trust_matrix} positions our approach in the evolution from classical planning through CSGM to post-condition verification.

\begin{table}[t]
\centering
\caption{Trust-Verification Matrix. Verification becomes mandatory as determinism decreases.}
\label{tab:trust_matrix}
\begin{tabular}{l l l l}
\toprule
\textbf{Framework} & \textbf{Primitive} & \textbf{Representation} & \textbf{Control} \\
\midrule
Classical PDDL & Deterministic & State-Transition & Open loop \\
CSGM / Goal-Net & Deterministic & Petri Net & Pre-task guard \\
\textbf{DSAP (Ours)} & \textbf{Stochastic ($\epsilon < 1$)} & \textbf{State-Transition} & \textbf{Post-task retry} \\
\bottomrule
\end{tabular}
\end{table}

\subsection{Promise Theory}

Promise Theory~\cite{Burgess2005,Burgess2014} models voluntary cooperation where autonomous agents issue promises of behavior rather than guarantees. The consumer bears sole responsibility for verifying promise fulfillment. Applied to our framework, the LLM is a ``promiser'' of artifacts; the guard function is the consumer's verification operator. This lens formalizes why post-condition verification is architecturally necessary rather than merely defensive.

\section{Formal Framework}
\label{sec:formal}

\begin{definition}[State Space Decomposition]
\label{def:state_space}
The system state space $\mathcal{S}$ decomposes into an observable workflow space and an opaque environment space:
\begin{equation}
\mathcal{S} = \Swork \times \Senv
\end{equation}
\begin{itemize}
    \item \textbf{Workflow State} ($\Swork$): The set of all status assignments to the guard functions:
    \begin{equation}
    \Swork = \{ \sigma \mid \sigma : \mathcal{G} \to \{\bot, \top, \bot_{\text{fatal}}\} \}
    \end{equation}
    where $\mathcal{G} = \{g_1, \ldots, g_n\}$ is the set of unique guard identifiers. Each guard maps to one of: $\bot$ (unsatisfied), $\top$ (satisfied), or $\bot_{\text{fatal}}$ (guard-determined unrecoverable failure).

    \item \textbf{Environment State} ($\Senv$): The Cartesian product of the artifact space and context space:
    \begin{equation}
    \Senv = \mathcal{A} \times \mathcal{C}
    \end{equation}
    A specific environment state is $s_{\text{env}} = \langle a, C \rangle$, where $a \in \mathcal{A}$ is the current artifact and $C \in \mathcal{C}$ is the cumulative context.
\end{itemize}
\end{definition}

The workflow state acts as a finite abstraction of execution progress. While guards return detailed feedback $\phi \in \Sigma^*$, this information resides in the opaque context $\mathcal{C}$. Only the tri-state verdict is retained in $\Swork$ ($|\Swork| = 3^{|\mathcal{G}|}$, finite), enabling the execution policy to distinguish retryable failures from guard-determined unrecoverable ones without inspecting the opaque context.

\begin{definition}[Scoped Context Composition]
\label{def:context}
The context $\mathcal{C}$ conditioning the generator and available to the guard composes three scopes:
\begin{equation}
C_{\text{total}} = \langle \mathcal{E}, C_{\text{local}}, H_{\text{feedback}} \rangle
\end{equation}
\begin{itemize}
    \item \textbf{Ambient Environment} ($\mathcal{E} = \langle \mathcal{R}, \Omega \rangle$):
    The versioned repository $\mathcal{R}$ providing read-only access to finalized artifacts, and global constraints $\Omega$.

    \item \textbf{Local Context} ($C_{\text{local}} = \langle \Psi \rangle$):
    The static specification for the current step. The current artifact $a_k$ is \emph{excluded} from $C_{\text{local}}$ to avoid a causality paradox: the artifact is the \emph{output} of the generator conditioned on $C_{\text{local}}$, not an input to it.

    \item \textbf{Feedback History} ($H_{\text{feedback}}$): Accumulated guard rejections for this step: $H = [(a_i, \phi_i), \ldots]$.
\end{itemize}
\end{definition}

\begin{remark}[Upstream Dependencies]
Prior step outputs are accessed via the Ambient Environment $\mathcal{E}$, specifically through the versioned repository $\mathcal{R}$. The workflow executor extracts relevant artifacts from $\mathcal{R}$ and passes them explicitly to the generator runtime.
\end{remark}

\begin{remark}[Artifact Space]
The artifact space $\mathcal{A}$ is formalized as an append-only versioned DAG $\mathcal{R}$ where nodes are artifact versions and edges are derivation steps. Every generative action creates a new node rather than overwriting, preserving the failure history for both auditability and future learning.
\end{remark}

\begin{definition}[Action Pair with Tri-State Guard]
\label{def:action_pair}
An action pair is a tuple $A = \langle \rho, \Agen, G \rangle$:
\begin{itemize}
    \item $\rho: \Swork \to \{0, 1\}$ is the \textbf{Precondition} (entry gate), determining applicability.
    \item $\Agen: \mathcal{C} \to \mathcal{A}$ is the \textbf{Generator}, producing an artifact from context.
    \item $G: \mathcal{A} \times \mathcal{C} \to \{\top, \bot_{\text{retry}}, \bot_{\text{fatal}}\} \times \Sigma^*$ is the \textbf{Guard} (exit gate), evaluating the artifact $a$ against context $C$ to yield a verdict and feedback. The inputs $a$ (the generation output) and $C$ (the conditioning context) are orthogonal: $a$ is the object under test; $C$ provides the validation criteria.
\end{itemize}
The tri-state guard distinguishes recoverable failures ($\bot_{\text{retry}}$: refine context and retry) from non-recoverable failures ($\bot_{\text{fatal}}$: escalate to human).
\end{definition}

\begin{definition}[System Dynamics]
\label{def:dynamics}
Given an action pair $A$ (Definition~\ref{def:action_pair}), the evolution of state $s_t = \langle s_{w,t}, s_{\text{env},t} \rangle$ proceeds in three phases:
\begin{enumerate}
    \item \textbf{Generation:} $a' \sim \Agen(C_t)$
    \item \textbf{Sensing:} $\langle v, \phi \rangle = G(a', C_t)$
    \item \textbf{State Update:} Determined by verdict $v$:
    \begin{itemize}
        \item If $v = \top$ (advance): $s_{t+1} = \langle T(s_w, \top), \langle a', \langle \Psi, \emptyset \rangle \rangle \rangle$
        \item If $v = \bot_{\text{retry}}$ (refine): $s_{t+1} = \langle s_w, \langle a', \langle \Psi, H_t \cup \{(a', \phi)\} \rangle \rangle \rangle$
        \item If $v = \bot_{\text{fatal}}$ (escalate): $s_{t+1} = \langle s_w[g_{\text{id}} \mapsto \bot_{\text{fatal}}], s_{\text{env},t} \rangle$; execution escalates to Level~3.
    \end{itemize}
\end{enumerate}
where $T(s_w, \top) = s_w[g_{\text{id}} \mapsto \top]$ updates the corresponding guard's truth value.
\end{definition}

\begin{definition}[Planning Problem]
\label{def:planning}
The planning problem is a tuple:
\begin{equation}
\mathcal{P} = \langle \Swork, \mathcal{A}, s_{w0}, C_{\text{init}}, \mathcal{S}_{\text{goal}}, \Rmax \rangle
\end{equation}
where $s_{w0}$ is the initial state (all guards $\bot$), $C_{\text{init}} = \langle \Psi, \emptyset \rangle$, and $\mathcal{S}_{\text{goal}} \subseteq \Swork$. A solution is a policy $\pi: \Swork \to \mathcal{A} \cup \{\text{wait}, \text{term}\}$ guaranteeing termination in $\mathcal{S}_{\text{goal}}$ subject to generator capability. In the current framework, plan synthesis is human-authored: the workflow topology is specified manually, and the planning problem reduces from search to reliability (Remark~\ref{rem:collapse}).
\end{definition}

\begin{algorithm}[t]
\caption{\textsc{Execute-Plan}($\pi, s_{w0}, \Rmax$)}
\label{alg:execute}
\begin{algorithmic}[1]
\STATE $\text{node} \gets \pi.\text{root}$; $s_w \gets s_{w0}$; $\text{ctx} \gets \textsc{InitContext}()$; $r \gets 0$
\WHILE{$\text{node} \neq \text{None}$}
    \STATE $A \gets \textsc{SelectAction}(\text{node}, s_w)$
    \IF{$r \geq \Rmax$}
        \RETURN $(\text{FAILURE})$
    \ENDIF
    \STATE $a \gets \textsc{Generate}(A.\Agen, \text{ctx})$ \COMMENT{Generation}
    \STATE $(v, \phi) \gets \textsc{EvalGuard}(A.G, a, \text{ctx})$ \COMMENT{Sensing}
    \IF{$v = \bot_{\text{retry}}$}
        \STATE $\text{ctx} \gets \textsc{Augment}(\text{ctx}, a, \phi)$; $r \gets r + 1$
    \ELSIF{$v = \bot_{\text{fatal}}$}
        \RETURN $(\text{ESCALATION\_REQUIRED})$
    \ELSE
        \STATE $s_w \gets \textsc{Transition}(s_w, v)$; $\text{ctx} \gets \textsc{ClearFeedback}(\text{ctx})$
        \STATE $\text{node} \gets \textsc{NextNode}(\text{node}, v)$; $r \gets 0$
    \ENDIF
\ENDWHILE
\RETURN $(\textsc{IsGoal}(s_w))$
\end{algorithmic}
\end{algorithm}

\begin{assumption}[Generator $\epsilon$-Capability]
\label{assum:capability}
For any valid specification context $C$, there exists $\epsilon > 0$ such that:
$P(\text{proj}_1(G(\Agen(C), C)) = \top) \geq \epsilon$.
\end{assumption}

\begin{proposition}[Asymptotic Soundness]
\label{prop:soundness}
Given an $\epsilon$-capable generator and finite plan $\pi$, the failure probability approaches~0:
\begin{equation}
P(\text{fail}) \leq (1 - \epsilon)^{\Rmax} \xrightarrow{\Rmax \to \infty} 0
\end{equation}
\end{proposition}

\begin{proof}
In the worst case (memoryless generator), each attempt succeeds independently with probability $\geq \epsilon$. The probability of $\Rmax$ consecutive failures at a single node is $(1{-}\epsilon)^{\Rmax}$. Since $(1{-}\epsilon) < 1$, this converges to 0. In practice, context refinement (Definition~\ref{def:context}) increases conditional success probability across retries, making this a conservative bound.
\end{proof}

\begin{corollary}[Reliability Bound]
\label{cor:reliability}
To achieve target global reliability $\delta$ across a $K$-step workflow:
\begin{equation}
\Rmax \geq \frac{\ln(1 - \delta^{1/K})}{\ln(1 - \epsilon)}
\end{equation}
\end{corollary}

\begin{remark}[Complexity Collapse]
\label{rem:collapse}
For fixed-topology (human-authored) workflows where the branching factor $B \approx 1$, the planning problem reduces from exponential search ($O(B^K)$) to a reliability problem dominated by the retry limit. The control complexity is $O(|\mathcal{S}_{\text{reach}}| \times \Rmax \times |\mathcal{G}|)$, linear in workflow length for sequential topologies.
\end{remark}

\begin{figure}[t]
    \centering
    \resizebox{0.95\textwidth}{!}{
    \begin{tikzpicture}[
        node distance=1.4cm and 1.8cm,
        auto,
        >=stealth,
        thick,
        font=\small,
        block/.style={rectangle, draw, fill=blue!5, text width=2.4cm, align=center, rounded corners, minimum height=2.8em},
        cloud/.style={ellipse, draw, fill=gray!10, text width=2.4cm, align=center, minimum height=2.8em, dashed},
        stack/.style={rectangle, draw, fill=white, text width=2.4cm, align=center, minimum height=2.8em, drop shadow={opacity=0.4}},
        decision/.style={diamond, draw, fill=yellow!10, text width=1.6cm, align=center, aspect=1.5},
        context/.style={rectangle, draw, fill=orange!5, text width=1.8cm, align=center, minimum height=2.8em}
    ]
    \node (start) [block] {Start State\\($s_w$)};
    \node (precond) [decision, right=of start] {Precondition\\$\rho(s_w)$};
    \node (ctx) [context, right=of precond] {Context\\$C$};
    \node (gen) [cloud, right=of ctx] {Generator\\$\Agen(C)$};
    \node (artifact) [stack, below=of gen, yshift=-0.3cm] {Artifact\\$a_k \in \mathcal{R}$};
    \draw[dashed, gray] ($(ctx.east)!0.5!(gen.west)$) coordinate (mid) -- +(0,-6.5) node[above, pos=-0.3, black, font=\footnotesize] {Control Boundary};
    \draw[dashed, gray] (mid) -- +(0,1.8);
    \draw[->] (start) -- (precond);
    \draw[->] (precond) -- node[above, pos=0.1, font=\footnotesize] {$\top$} (ctx);
    \draw[->] (ctx) -- (gen);
    \draw[->] (gen) -- node[right, font=\footnotesize] {Produces} (artifact);
    \node (guard) [decision, below=of precond, yshift=-2.2cm] {Guard\\$G(a, C)$};
    \draw[->] (artifact.south) |- node[near start, right, font=\footnotesize] {Senses $a_k$} (guard.east);
    \draw[->, dotted] (ctx.south) -- ++(0,-0.8) -| (guard.north);
    \node (success) [block, left=of guard] {Next State\\($s_{w+1}$)};
    \draw[->] (guard) -- node[above, font=\footnotesize] {$\top$} (success);
    \draw[->, red, very thick, rounded corners]
        (guard.south) -- node[right, font=\footnotesize, pos=0.1] {$\bot_{\text{retry}}$} ++(0,-0.7)
        -| ($(ctx.south) + (0.4, -0.4)$)
        -- node[right, font=\footnotesize, pos=0.0] {Refine $C$} (ctx.south);
    \node[above=0.15cm of gen, font=\bfseries\footnotesize] {Environment State};
    \node[above=0.15cm of precond, xshift=1.2cm, font=\bfseries\footnotesize] {Workflow State};
    \end{tikzpicture}
    }
    \caption{The Atomic Action Pair. The control boundary separates the observable workflow (left) from the opaque environment (right). The red loop shows context refinement: $s_w$ remains invariant while guard feedback updates the generative context $C$.}
    \label{fig:atomic_pair}
\end{figure}

\section{Recovery Architecture}
\label{sec:recovery}

\subsection{The Stagnation Problem}

Algorithm~\ref{alg:execute} handles failures through Level~1 recovery: context refinement within a single step. When the root cause lies upstream---e.g., a code patch repeatedly fails because the \emph{analysis} step identified the wrong file---local retries exhaust $\Rmax$ without progress. Naive exhaustive backtracking across all $K$ steps with budget $R$ per step risks $O(R^K)$ exploration, which is intractable for non-trivial workflows.

We introduce a three-level recovery hierarchy:
\begin{description}
    \item[Level 1: Context Refinement.] Retry within the same step, augmenting context with guard feedback (Algorithm~\ref{alg:execute}, line~10).
    \item[Level 2: Informed Backtracking.] Detect stagnation, cascade-invalidate dependent steps, inject failure context into upstream steps, and re-execute (this section).
    \item[Level 3: Human Escalation.] When automated recovery budgets ($\Rmax$, $\emax$) are exhausted, or when the guard returns $\bot_{\text{fatal}}$.
\end{description}
Escalation routing---which upstream step to revisit---is currently human-configured in workflow specifications. The mechanism is a tool for execution recovery, not an automated planner.

\begin{figure}[t]
    \centering
    \begin{tikzpicture}[
        >=stealth,
        font=\small,
        level/.style={rectangle, draw, rounded corners, minimum width=3.2cm, minimum height=2em, align=center},
        arrow/.style={->, thick}
    ]
    \node[level, fill=green!10] (l1) at (0,0) {Level 1: Context Refinement};
    \node[level, fill=yellow!15] (l2) at (0,-1.8) {Level 2: Informed Backtracking};
    \node[level, fill=red!10] (l3) at (0,-3.6) {Level 3: Human Escalation};

    \node[right=0.4cm of l1, font=\footnotesize, text width=4.5cm] {Retry within step; augment context with guard feedback $\phi$};
    \node[right=0.4cm of l2, font=\footnotesize, text width=4.5cm] {Stagnation $\to$ cascade invalidate $\to$ inject failure context $\to$ upstream re-execute};
    \node[right=0.4cm of l3, font=\footnotesize, text width=4.5cm] {$\emax$ exhausted or $\bot_{\text{fatal}}$; workflow pauses for human};

    \draw[arrow] (l1.south) -- node[left, font=\footnotesize] {$\rpatience$ similar failures} (l2.north);
    \draw[arrow] (l2.south) -- node[left, font=\footnotesize] {$\emax$ exhausted} (l3.north);
    \end{tikzpicture}
    \caption{Three-level recovery hierarchy. Each level has an explicit budget; escalation occurs when the current level's budget is exhausted.}
    \label{fig:recovery}
\end{figure}

\subsection{Formal Definitions}

\begin{definition}[Stagnation Detection]
\label{def:stagnation}
Let $H_k = [(a_1, \phi_1), \ldots, (a_k, \phi_k)]$ be the feedback history for a step. Stagnation is detected when $\rpatience$ consecutive failures exhibit repetitive patterns:
\begin{equation}
\textsc{Stagnant}(H_k, \rpatience) \equiv |H_k| \geq \rpatience \;\wedge\; \big(\textsc{Repetition}(H_k, \rpatience) \;\vee\; \textsc{Similarity}(H_k, \rpatience, \theta)\big)
\end{equation}
where $\textsc{Repetition}$ checks for identical error signatures among the last $\rpatience$ entries, and $\textsc{Similarity}$ checks whether all pairwise feedback similarities exceed threshold $\theta$ (we use $\theta = 0.7$). The constraint $1 < \rpatience < \Rmax$ ensures at least one retry before escalation while leaving room for Level~2 to act before $\Rmax$ exhaustion.

For composite guards (multiple sub-guards), stagnation detection operates per sub-guard stream independently, catching oscillation patterns where alternating failures across sub-guards mask stagnation in the global sequence.
\end{definition}

\begin{definition}[Cascade Invalidation]
\label{def:cascade}
When backtracking to step $j$, the workflow invalidates step $j$ and all its transitive dependents:
\begin{equation}
\textsc{Invalidate}(s_w, j) = s_w\big[g_j \mapsto \bot,\; g_m \mapsto \bot \;\forall m \in \textsc{Dependents}^+(j)\big]
\end{equation}
where $\textsc{Dependents}^+(j) = \{m \mid j \in \text{requires}^+(m)\}$ is computed via breadth-first traversal of the dependency DAG. Invalidation resets any guard state---including $\bot_{\text{fatal}}$---to $\bot$, since the new upstream context may avoid the conditions that caused the fatal determination. Cached artifacts for invalidated steps are evicted, ensuring re-execution produces fresh outputs consistent with the new upstream state.
\end{definition}

\begin{definition}[Context Injection]
\label{def:injection}
When escalating from step $k$ to upstream step $j$, a failure summary $\sigma_k$ is injected into step $j$'s specification context:
\begin{equation}
C_j' = C_j \oplus \sigma_k
\end{equation}
The failure summary $\sigma_k$ is distinct from step $j$'s own feedback history $H_j$. It carries cross-step causal information: the error patterns observed downstream, the approaches already tried, and a constraint directing the upstream generator to produce output that avoids the identified failure modes.
\end{definition}

\subsection{Execution with Recovery}

Algorithm~\ref{alg:recovery} extends Algorithm~\ref{alg:execute} with Level~2 and Level~3 recovery.

\begin{algorithm}[t]
\caption{\textsc{Execute-Plan-With-Recovery}($\pi, s_{w0}, \Rmax, \rpatience, \emax$)}
\label{alg:recovery}
\begin{algorithmic}[1]
\STATE Initialize $s_w, \text{ctx}, \text{escalation\_count}$ as in Alg.~\ref{alg:execute}
\WHILE{$\neg\textsc{IsGoal}(s_w)$}
    \STATE $\text{step} \gets \textsc{FindApplicable}(s_w)$
    \IF{$\text{step} = \text{None}$}
        \RETURN FAILURE
    \ENDIF
    \STATE $\text{ctx} \gets \textsc{InjectContext}(\text{ctx}, \text{step})$ \COMMENT{Apply any pending $\sigma$}
    \STATE $(\text{result}, H) \gets \textsc{ExecuteStep}(\text{step}, \text{ctx}, \Rmax)$ \COMMENT{Alg.~\ref{alg:execute} inner loop}
    \IF{result $= \top$}
        \STATE $s_w \gets \textsc{Transition}(s_w, \text{step})$ \COMMENT{Advance}
    \ELSIF{result $= \bot_{\text{fatal}}$}
        \STATE $s_w \gets s_w[g_{\text{step}} \mapsto \bot_{\text{fatal}}]$ \COMMENT{Record fatal}
        \RETURN ESCALATION\_REQUIRED \COMMENT{Level 3: guard-determined}
    \ELSIF{$\textsc{Stagnant}(H, \rpatience)$ \textbf{and} $\text{targets} \neq \emptyset$ \textbf{and} $\text{esc\_count} < \emax$}
        \STATE $\sigma \gets \textsc{FailureSummary}(H)$ \COMMENT{Definition~\ref{def:injection}}
        \FOR{$j \in \text{targets}$}
            \STATE $s_w \gets \textsc{Invalidate}(s_w, j)$ \COMMENT{Definition~\ref{def:cascade}}
            \STATE $\textsc{StoreInjection}(j, \sigma)$
        \ENDFOR
        \STATE $\text{esc\_count} \gets \text{esc\_count} + 1$; \textbf{continue} \COMMENT{Re-run from earliest}
    \ELSE
        \RETURN ESCALATION\_REQUIRED \COMMENT{Level 3: budget exhausted}
    \ENDIF
\ENDWHILE
\RETURN SUCCESS
\end{algorithmic}
\end{algorithm}

\subsection{Termination Properties}

\begin{proposition}[Termination under Recovery]
\label{prop:termination}
Algorithm~\ref{alg:recovery} terminates within $O(K \times \Rmax \times \emax)$ generation attempts, where $K$ is the number of workflow steps.
\end{proposition}

\begin{proof}
Each step permits at most $\Rmax$ retries before either succeeding or triggering escalation. Each escalation invalidates upstream steps, which are then re-executed with at most $\Rmax$ retries each. With $\emax$ escalation attempts per step and $K$ steps, the total generation attempts are bounded by $K \times \Rmax \times (\emax + 1)$.
\end{proof}

\begin{proposition}[$O(R^K)$ Prevention]
\label{prop:ork}
The recovery hierarchy prevents naive $O(R^K)$ backtracking. Stagnation detection (Definition~\ref{def:stagnation}) triggers Level~2 escalation after $\rpatience$ similar failures, rather than exhausting all $\Rmax$ retries before backtracking. The $\emax$ bound further caps the total escalation budget, yielding worst-case $O(K \times \Rmax \times \emax)$ instead of $O(\Rmax^K)$.
\end{proposition}

\begin{proof}
In naive backtracking, each of $K$ steps could exhaust $\Rmax$ retries for every combination of upstream outputs, yielding $\Rmax^K$. With stagnation detection, escalation is triggered after at most $\rpatience < \Rmax$ similar failures. Each escalation increments a counter bounded by $\emax$. Since escalation counts are per-step and bounded, the total attempts across all steps sum to at most $K \times \Rmax \times (\emax + 1) \in O(K \times \Rmax \times \emax)$.
\end{proof}

\section{Experimental Validation}
\label{sec:experiments}

\subsection{Setup}

Experiments evaluate the DSAP framework on three diagnostic probes designed to isolate specific failure modes. Each probe targets a different ``implementation gap'' between conceptual knowledge and correct generation:

\begin{itemize}
    \item \textbf{LRU Cache:} High prior knowledge, standard implementation. Guards prevent stochastic \emph{drift} in well-known patterns.
    \item \textbf{Template Engine:} High conceptual prior, novel implementation. Guards provide \emph{structural feedback} for parser synthesis.
    \item \textbf{Password Validator:} High conceptual prior, \emph{calculation gap}. Rules require mathematical operations (prime number computation) difficult for token-prediction models.
\end{itemize}

13 models from 6 families (1.3B--15B parameters) were evaluated locally via Ollama, with temperature $T{=}0.7$ to stress-test the framework against high-variance generation. Each model--task pair was tested across 50 independent trials with context isolation between trials. Two configurations were compared: \textbf{Baseline} (single attempt, $k{=}1$) and \textbf{Guarded} ($\Rmax{=}3$).

A model is \emph{qualified} if it consistently produces parsable output in zero-shot evaluation. Models failing this threshold ($\epsilon \approx 0$) are excluded from gain analysis, as including them would conflate architectural failure with model incapacity.

\subsection{DSAP Effectiveness: Diagnostic Probes}

Table~\ref{tab:results} presents condensed results across all three probes. Statistical significance was assessed using Fisher's exact test, with effect sizes reported as Cohen's $h$.

\begin{table}[t]
\centering
\small
\caption{Diagnostic probe results (top models by reliability gain). Statistical significance: *** $p{<}0.001$, ** $p{<}0.01$, * $p{<}0.05$ (Fisher's exact test). 50 trials per cell.}
\label{tab:results}
\begin{tabular}{@{}l l c c c c@{}}
\toprule
\textbf{Task} & \textbf{Model} & \textbf{Base} & \textbf{Guarded} & \textbf{Gain ($\Delta$)} & \textbf{Avg.\ Retries} \\
\midrule
\multirow{3}{*}{\textbf{Password}} & StarCoder2 (15B) & 0\% & 66\% & \textbf{+66}*** & 0.84 \\
 & DeepSeek-Coder (6.7B) & 50\% & 96\% & \textbf{+46}*** & 0.72 \\
 & Granite-Code (3B) & 36\% & 80\% & \textbf{+44}*** & 1.46 \\
\midrule
\multirow{3}{*}{\textbf{LRU Cache}} & DeepSeek-Coder (6.7B) & 48\% & 98\% & \textbf{+50}*** & 0.76 \\
 & Granite-Code (8B) & 60\% & 98\% & \textbf{+38}*** & 0.52 \\
 & Yi-Coder (1.5B) & 62\% & 98\% & \textbf{+36}*** & 0.76 \\
\midrule
\multirow{3}{*}{\textbf{Template}} & Yi-Coder (9B) & 56\% & 98\% & \textbf{+42}*** & 0.92 \\
 & StarCoder2 (15B) & 60\% & 100\% & \textbf{+40}*** & 0.32 \\
 & Qwen2.5-Coder (3B) & 8\% & 42\% & \textbf{+34}*** & 2.72 \\
\bottomrule
\end{tabular}
\end{table}

\textbf{Password Validator (Calculation Gap):} The largest effect size was observed for StarCoder2~(15B): from 0\% to 66\% ($\Delta{=}{+}66$pp, Cohen's $h{=}1.90$). This model understands the \emph{structure} of prime-checking but cannot reliably \emph{compute} primes zero-shot. Guards bootstrap reasoning by providing concrete counterexamples on each failure.

\textbf{LRU Cache (Drift Prevention):} 11 of 13 models achieved $\geq$98\% guarded success. DeepSeek-Coder~(6.7B) showed the largest gain ($\Delta{=}{+}50$pp, Cohen's $h{=}1.33$), demonstrating that guards effectively close the reliability gap for mid-capability models on well-known patterns.

\textbf{Template Engine (Structural Gap):} Yi-Coder~(9B) improved from 56\% to 98\% ($\Delta{=}{+}42$pp, Cohen's $h{=}1.17$). Sub-3B models showed negligible improvement ($\epsilon \approx 0$), establishing a clear capability threshold below which guards cannot compensate.

\textbf{Efficiency:} The framework achieves convergence at 1.2--2.1$\times$ baseline cost for qualified models, compared to a fixed 5.0$\times$ for Pass@5 sampling. Model qualification ($\epsilon > 0$) is task-specific: Phi4-Mini~(3.8B) is qualified for LRU (60\% baseline) but unqualified for Password (0\%).

\subsection{Recovery Mechanism Behavior}

To validate the recovery hierarchy (C2), the framework was applied to 99 instance-arm pairs from SWE-Bench Pro~\cite{jimenez2024swebench} (test split, Python, 50 unique instances) using Qwen3-Coder-Next via OpenRouter. Two TDD workflow variants were evaluated (\texttt{tdd\_verified}, \texttt{tdd\_behavior}). Each variant executes a three-step pipeline: analysis $\to$ \texttt{gen\_test} $\to$ \texttt{gen\_patch}, with $\Rmax{=}6$ and composite guards enforcing per-guard escalation routing. Step-level parameters: \texttt{ap\_gen\_test} ($\rpatience{=}2$, $\emax{=}1$, escalates to analysis); \texttt{ap\_gen\_patch} ($\rpatience{=}3$, $\emax{=}2$, escalates to analysis and \texttt{gen\_test}). Guards include PatchGuard, LintGuard, TestGreenGuard, TestRedGuard, and FullEvalGuard, each with independently configured escalation targets. Note: the SWE-Bench experiments evaluate recovery mechanism \emph{behavior} (escalation patterns, injection effectiveness, stagnation detection); Docker-based SWE-Bench evaluation was not performed, as the primary goal is architectural validation rather than benchmark competition.

\begin{table}[t]
\centering
\small
\caption{Recovery mechanism metrics from SWE-Bench Pro (99 instance-arm pairs, Qwen3-Coder-Next, two TDD workflow variants). No Docker evaluation was performed; metrics reflect guard-level execution behavior.}
\label{tab:recovery}
\begin{tabular}{@{}l r@{}}
\toprule
\textbf{Metric} & \textbf{Value} \\
\midrule
Instances attempted & 99 \\
Steps resolved at Level 1 (retry only) & 56/278 (20.1\%) \\
Stagnation events (Level 2 triggered) & 71 \\
Total escalation cycles & 190 \\
Mean cascade depth per escalation & 1.7 steps \\
Context injection events & 71 \\
Upstream output changed after injection & 71/71 (100\%) \\
\midrule
\multicolumn{2}{@{}l}{\textit{Backtracking Recovery Rate}} \\
\quad \texttt{ap\_gen\_test} & 6/16 (37.5\%) \\
\quad \texttt{ap\_gen\_patch} & 0/55 (0.0\%) \\
\midrule
\multicolumn{2}{@{}l}{\textit{Terminal State Distribution}} \\
\quad All attempted steps passed guards & 4 (4.0\%) \\
\quad $\emax$ exhausted & 10 (10.1\%) \\
\quad $\Rmax$ exhausted (no escalation) & 85 (85.9\%) \\
\bottomrule
\end{tabular}
\end{table}

Table~\ref{tab:recovery} summarizes the recovery metrics across three dimensions.

\textbf{Level~1 (Retry) Behavior.} Only 20.1\% of steps resolved through retry alone. The dominant terminal state---85 of 99 instances (85.9\%) exhausting $\Rmax$ without escalation ever triggering---indicates that the LLM produces sufficiently different failing outputs on each attempt to avoid stagnation detection, consuming the retry budget without progress.

\textbf{Level~2 (Escalation) Mechanical Correctness.} Stagnation detection triggered 71 escalation events across 99 instances. Context injection was mechanically effective in every case: upstream generators produced different output after receiving the injected failure summary (71/71, 100\%). Guard-specific escalation routing operated as specified: TestGreenGuard escalated to both analysis and \texttt{gen\_test}, while TestRedGuard escalated to analysis only (Table~\ref{tab:guards}).

\textbf{Backtracking Recovery Rate.} The critical finding is a sharp step-specific asymmetry. Escalation recovered \texttt{ap\_gen\_test} in 6 of 16 stagnation events (37.5\%)---when test generation stagnated and escalated to analysis, the regenerated analysis enabled test generation to succeed. In contrast, escalation recovered \texttt{ap\_gen\_patch} in 0 of 55 events (0.0\%)---despite upstream output always changing, the regenerated analysis and tests never guided the patch generator to a passing solution. All four workflow completions (4.0\%) were achieved at Level~1 only; no escalation-assisted instance achieved full workflow completion. Notably, two of the four completions only reached the test generation step---they did not attempt patch generation---indicating that ``completion'' reflects guard satisfaction on attempted steps, not full pipeline execution. No instance in either arm produced a viable patch. The bottleneck is \texttt{ap\_gen\_patch}, which accounts for 80 of 95 failed instances, with TestGreenGuard as the most common failing guard (44 final failures). All 10 $\emax$-exhausted instances involved TestRedGuard on \texttt{ap\_gen\_test}---test generation is the only step that exhausts its escalation budget, precisely because it is the only step where escalation has nonzero recovery probability.

The recovery hierarchy is mechanically sound \emph{and} partially effective---it demonstrably rescues test generation 37.5\% of the time. The sharp asymmetry (37.5\% vs.\ 0.0\%) reveals a complexity threshold: escalation helps ``easier'' generation tasks (tests) but cannot bridge the gap for ``harder'' ones (patches that must pass those tests). This is the empirical basis for the boundary finding in Section~\ref{sec:boundary}.

\begin{table}[t]
\centering
\small
\caption{Guard-specific failure and escalation patterns across 99 SWE-Bench instances. Stagnation triggers indicate how often each guard's failure stream activated Level~2 escalation.}
\label{tab:guards}
\begin{tabular}{@{}l r r l@{}}
\toprule
\textbf{Guard} & \textbf{Failures} & \textbf{Stagnation} & \textbf{Escalation Targets} \\
\midrule
TestGreenGuard & 579 & 35 & analysis, gen\_test \\
PatchGuard & 207 & 9 & analysis, gen\_test \\
AnalysisGuard & 105 & 0 & --- \\
LintGuard & 92 & 7 & analysis, gen\_test \\
TestRedGuard & 77 & 14 & analysis \\
GeneratorValidation & 51 & 5 & analysis, gen\_test \\
FullEvalGuard & 34 & 0 & --- \\
TestSyntaxGuard & 6 & 1 & analysis \\
\bottomrule
\end{tabular}
\end{table}

\subsection{Boundary Finding: Execution vs.\ Planning}
\label{sec:boundary}

Of 99 instance-arm pairs, only 4 (4.0\%) completed all attempted workflow steps with guard satisfaction---and none produced a final patch. This establishes a clear boundary between execution architecture and plan synthesis. The four completions were achieved at Level~1 (retry only)---the recovery hierarchy's higher levels never contributed to full workflow completion, despite demonstrably recovering intermediate steps.

The step-specific recovery asymmetry provides the sharpest evidence for a complexity threshold. Escalation recovers \texttt{ap\_gen\_test} 37.5\% of the time: when test generation stagnates, the regenerated upstream analysis is sufficient to unblock it. But escalation recovers \texttt{ap\_gen\_patch} 0\% of the time: even with 100\% upstream output change after context injection, the regenerated analysis and tests never guide the patch generator to functional correctness. The 37.5\% test recovery rate is itself a positive result---it validates that the escalation mechanism \emph{can} work when the generation task is tractable---but the 0\% patch recovery rate reveals where execution recovery reaches its limit.

Five failure modes were identified, reframed around this asymmetry:

\begin{enumerate}
    \item \textbf{Search-string hallucination:} The patch step generates search strings that do not exist in the target file. Escalation regenerates the analysis, but the new analysis does not prevent the patch generator from hallucinating different incorrect search strings.
    \item \textbf{Wrong escalation axis:} Backtracking re-runs analysis (localization), but the localization was correct---the fix \emph{strategy} was wrong. The workflow has no mechanism to select a different fix approach, only to retry the same approach with different upstream context.
    \item \textbf{Context overflow:} Accumulated feedback and injected context exceed the model's effective window, degrading rather than improving generation quality across escalation cycles.
    \item \textbf{Insufficient escalation depth:} 85/99 instances exhaust $\Rmax$ without stagnation detection triggering---the LLM produces sufficiently varied failing outputs to evade stagnation detection, consuming the retry budget without ever escalating.
    \item \textbf{Framework-specific failures:} The static workflow assumes a generic patch strategy; framework-specific code patterns require specialized decomposition that no amount of escalation within the current workflow can provide.
\end{enumerate}

All five failure modes trace to a single root cause: \emph{static workflow selection with no instance awareness}. The human-authored workflow specifies a fixed analysis $\to$ test $\to$ patch pipeline. Execution recovery can retry and backtrack within this pipeline, but it cannot select a different pipeline. The binding constraint is not the escalation mechanism---which demonstrably works for tractable subtasks---but the inherent difficulty of code patch generation under a fixed decomposition strategy. This is precisely the planning problem: given an issue description, synthesize an appropriate workflow. The recovery hierarchy is necessary---without it, the system exhausts $\Rmax$ on the wrong approach without ever attempting correction---but not sufficient for autonomous resolution of diverse software engineering tasks.

\subsection{Analysis}

Guard effectiveness is task-specific and model-specific. The framework provides guarantees within a valid plan; plan synthesis is orthogonal. Three key findings emerge:

\begin{enumerate}
    \item \textbf{Architectural intervention dominates parameter scale.} Mid-sized models (3B--9B) with guards frequently outperform larger models without guards.
    \item \textbf{Qualification is per-task.} Guard-based systems should assess model capability per-task rather than assuming uniform competence.
    \item \textbf{Execution recovery establishes failure taxonomy.} The SWE-Bench results provide a structured decomposition of where execution architecture ends and planning begins.
\end{enumerate}

\section{Discussion}

\textbf{Graceful Degradation.} The three-level hierarchy provides graduated recovery: most failures resolve at Level~1 (cheap), some require Level~2 (moderate), and only intractable cases reach Level~3 (human). This mirrors established engineering practices where automated recovery handles common cases and human expertise addresses edge cases.

\textbf{The Execution--Planning Boundary.} The SWE-Bench results sharpen a distinction often blurred in LLM agent literature. The DSAP is an execution atom; the recovery hierarchy is an execution control mechanism. Neither substitutes for plan synthesis---selecting \emph{which} workflow to execute for a given problem instance (Definition~\ref{def:planning}). This is the next research question.

\textbf{Credit Assignment.} By enforcing immediate verification after each generation attempt (Definition~\ref{def:dynamics}), the architecture naturally produces dense, attributable reward signals~\cite{sutton2018reinforcement}. Each (artifact, guard\_result) pair is a labeled training example with zero temporal distance. This structural property may support future integration with reinforcement learning for generator improvement.

\textbf{Future Work.} HTN-style hierarchical task network planning could automate workflow synthesis, selecting among parameterized workflow templates based on problem classification. Dynamic guard selection---choosing verification strategies at runtime based on observed failure patterns---would further reduce the human specification burden.

\section{Conclusion}

We have formalized the Dual-State Action Pair, an execution primitive that couples stochastic LLM generation with deterministic post-condition verification. The framework decomposes system state into finite, observable workflow state and infinite, opaque environment state, enabling convergence guarantees for $\epsilon$-capable generators.

The three-level recovery hierarchy---context refinement, informed backtracking, and human escalation---prevents $O(R^K)$ retry explosion (Proposition~\ref{prop:termination}) while maintaining bounded worst-case complexity of $O(K \times \Rmax \times \emax)$. Escalation routing is human-configured, providing a formally grounded mechanism without overclaiming automated planning capabilities.

Experimental validation demonstrates reliability gains of up to 66 percentage points across 13 LLMs on diagnostic probes, at modest computational overhead. Recovery mechanism evaluation on 99 SWE-Bench Pro instance-arm pairs validates the hierarchy's mechanical correctness---100\% context injection effectiveness across 71 escalation events, with 37.5\% recovery rate for test generation---while establishing through 0\% patch production and 0\% patch generation recovery that execution architecture is \emph{necessary but not sufficient} for autonomous software engineering agents. Plan synthesis remains the open problem.

\section*{Acknowledgements}

This work was conducted as part of ongoing independent Masters research. The formal definitions and algorithms in this paper are sufficient for independent reimplementation. The reference implementation and experimental data are available from the author upon request.

\bibliographystyle{plain}
\bibliography{refs}

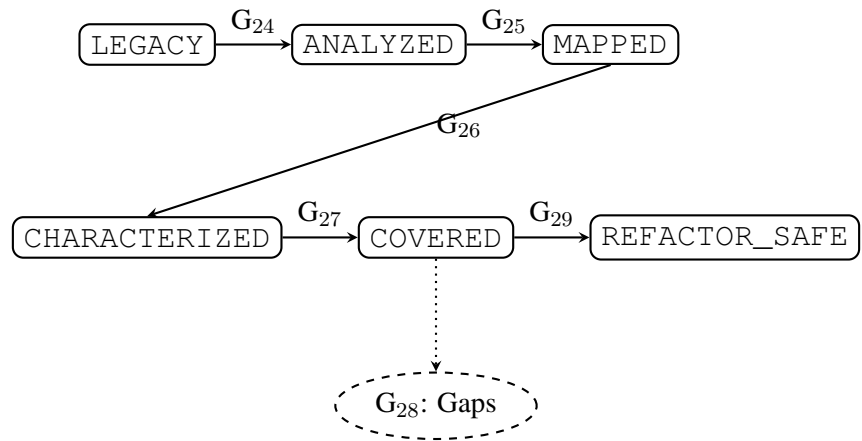
\end{document}